\documentclass{article}

\usepackage{microtype}
\usepackage{graphicx}
\usepackage{subcaption}
\usepackage{booktabs}
\usepackage{hyperref}

\usepackage[preprint]{icml2026}
\usepackage{amsmath}
\usepackage{amssymb}
\usepackage{mathtools}
\usepackage{amsthm}
\usepackage[capitalize,noabbrev]{cleveref}

\usepackage[table]{xcolor}
\usepackage{colortbl}
\newif\iftablecolor
\tablecolortrue      

\newcommand{\myrowcolor}[1]{%
    \iftablecolor
        \rowcolor{#1}%
    \fi
}

\definecolor{PurpleGroup}{RGB}{165,140,200}
\definecolor{PurpleLight}{RGB}{235,230,245}
\definecolor{OrangeGroup}{RGB}{231,111,81}
\definecolor{OrangeLight}{RGB}{250,220,210}

\definecolor{BlueLight}{RGB}{235,243,250}

\newcommand{\ps}{p_{\mathrm{S}}}
\newcommand{\pt}{p_{\mathrm{T}}}
\newcommand{\zs}{z_y}
\newcommand{\sg}{\operatorname{sg}}

\icmltitlerunning{\smash{Bernoulli OPD}}

\begin{document}

\twocolumn[
  \icmltitle{Bernoulli On-Policy Distillation: Don’t Ignore High-Confidence Student Tokens}

  \begin{icmlauthorlist}
    \icmlauthor{Mingyu Cao}{surrey}
    \icmlauthor{Yaokai Jin}{ucl}
    \icmlauthor{Shiwei Liu}{ellis,mpi,tueai}
    \icmlauthor{Lu Yin}{surrey}
  \end{icmlauthorlist}

  \icmlaffiliation{surrey}{
    University of Surrey
  }
  \icmlaffiliation{ucl}{
    University College London
  }
  \icmlaffiliation{ellis}{
    ELLIS Institute Tübingen
  }
  \icmlaffiliation{mpi}{
    Max Planck Institute for Intelligent Systems
  }
  \icmlaffiliation{tueai}{
    Tübingen AI Center
  }

  \icmlcorrespondingauthor{Lu Yin}{l.yin@surrey.ac.uk}

  \icmlkeywords{
    on-policy distillation,
    knowledge distillation,
    language models,
    optimization geometry
  }

  \vskip 0.3in
]

\printAffiliationsAndNotice{}

\begin{abstract}
On-policy distillation (OPD) trains a student language model on its own trajectories using token-level supervision from a stronger teacher.
We show that standard sampled reverse-KL OPD systematically suppresses updates to high-confidence tokens, such that their weak gradient contribution does not necessarily imply weak or uninformative teacher supervision.
We introduce \textbf{Bernoulli On-Policy Distillation (Bernoulli OPD)}, which directly matches the teacher and student probabilities for that token.
This yields a bounded probability space correction that preserves teacher--student disagreement while avoiding the additional confidence-dependent suppression induced by vanilla OPD.
Fixed-rollout analyses show that Bernoulli OPD substantially restores optimization signal to high-confidence tokens while remaining sensitive to teacher–student disagreement.
We further analyze the resulting model behavior through confidence distributions, answer calibration, policy movement, and training dynamics, showing that Bernoulli OPD's gains cannot be explained simply by longer reasoning, larger optimization steps, or more uniform imitation of the teacher. Code is available at https://github.com/duterscmy/Bernoulli-OPD.
\end{abstract}

\section{Introduction}

\begin{figure*}[t]
    \centering
    \includegraphics[width=0.85\textwidth]
    {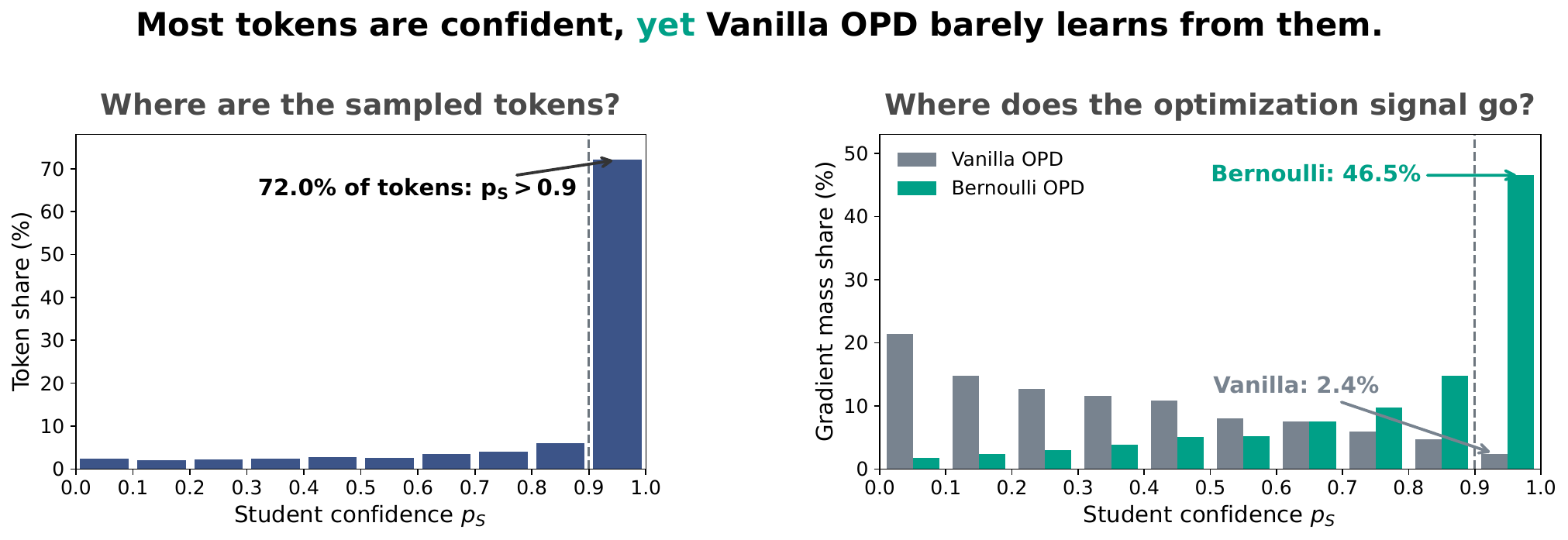}
\caption{
\textbf{Token distribution and gradient allocation across student-confidence regions.}
In our primary setting, about $72\%$ of sampled tokens are high-confidence with $p_S>0.9$.
However, this region receives only about $2.4\%$ of the total gradient mass under vanilla OPD, compared with about $46.5\%$ under Bernoulli OPD.
The contrast between token frequency and gradient allocation highlights the strong confidence-dependent suppression induced by vanilla sampled reverse-KL OPD.
}
    \label{fig:vanilla-gradient-distribution}
\end{figure*}

On-policy distillation (OPD) trains a student on its own generated
trajectories while using a stronger teacher to provide token-level
supervision on the visited states
\citep{agarwal2024onpolicydistillationlanguagemodels,gu2026minillmonpolicydistillationlarge,song2026surveyonpolicydistillationlarge}.
By learning directly on student-generated trajectories, OPD reduces the
distribution mismatch inherent to offline distillation and has shown
strong potential for improving language-model reasoning.

Despite its effectiveness, the optimization behavior of sampled-token
OPD remains poorly understood.
In the commonly used sampled reverse-KL objective, the gradient assigned
to a sampled token depends not only on teacher--student disagreement,
but also strongly on the student's own confidence.
As the student becomes more confident in a sampled token, its gradient
is increasingly suppressed, even when the teacher still assigns a
meaningfully different probability to that token.
This introduces \textbf{a confidence-dependent bias} into how the objective
allocates optimization signal.

Figure~\ref{fig:vanilla-gradient-distribution} illustrates this effect.
High-confidence tokens constitute a large fraction of sampled tokens in
student-generated trajectories, yet receive only a small fraction of
the total gradient mass under vanilla sampled OPD.
Importantly, weak gradient contribution does not necessarily imply that
these tokens contain little useful teacher supervision.
When we train exclusively on highly confident sampled tokens, the
student still improves substantially throughout training, although it
remains weaker than training on all sampled tokens.
This suggests that the limited contribution of high-confidence tokens
under vanilla OPD is at least partly induced by the geometry of the
objective itself rather than by a lack of useful learning signal.

These observations motivate a simple question:
\textbf{can we effectively learn from teacher corrections even when the
student is already confident in its prediction?}
To address this issue, we propose \textbf{Bernoulli On-Policy
Distillation (Bernoulli OPD)}.
Instead of using the sampled reverse-KL log ratio objective, Bernoulli
OPD treats each sampled token as a simple binary prediction and directly
matches the teacher and student probabilities for that token.
This gives a bounded update that follows the difference between the
teacher and student probabilities, without further weakening the
training signal on high-confidence tokens.

Our experiments support both the proposed mechanism and its practical
utility.
On identical student rollouts, Bernoulli OPD allocates substantially
more optimization signal to high-confidence tokens while remaining
sensitive to teacher--student disagreement.
In our primary setting, Bernoulli OPD improves the five-benchmark
average from 44.38 to 51.76 over vanilla OPD, and this advantage is
reproduced across three independent training seeds
(Appendix~\ref{app:seed_robustness}). More broadly, Bernoulli OPD consistently improves over vanilla OPD
across four teacher--student settings, covering both Qwen3 and
DeepSeek-based students, different types of teacher models, and both non-thinking and thinking generation modes.

\paragraph{Our contributions are summarized as follows.}
\begin{itemize}
    \item We show that sampled reverse-KL OPD gives very little
    optimization signal to high-confidence tokens, even though these
    tokens are common in student rollouts and can still provide useful
    learning signal.

    \item We propose Bernoulli OPD, a simple objective that directly
    matches teacher and student token probabilities and avoids
    unnecessarily weakening updates on high-confidence tokens.

    \item Across four teacher--student settings, Bernoulli OPD
consistently improves reasoning performance over vanilla OPD.
We further analyze its behavior through confidence distributions,
answer-level calibration, policy movement, and training dynamics.
\end{itemize}

\section{Related Work}
\label{sec:related}

\paragraph{Off-policy and on-policy distillation.}
Classical knowledge distillation learns from a fixed data distribution \citep{hinton2015distillingknowledgeneuralnetwork}, while sequence-level distillation often uses teacher-generated responses \citep{kim2016sequencelevelknowledgedistillation}.
On-policy distillation instead trains on student generated trajectories, reducing the mismatch between training and inference states \citep{agarwal2024onpolicydistillationlanguagemodels,gu2026minillmonpolicydistillationlarge}.
Existing OPD methods can be broadly divided into full vocabulary \citep{jin2026entropyawareonpolicydistillationlanguage}, Top-$k$ \citep{meng2026unbiasedtopkestimationonpolicy}, and sampled token variants \citep{gu2026minillmonpolicydistillationlarge}.
Full vocabulary methods use dense teacher distributions, Top-$k$ methods retain only a restricted support, and sampled-based methods use student sampled tokens as Monte Carlo estimates.
The latter is particularly attractive for long horizon reasoning research because of its lower memory and communication cost.
Our work also focuses on this sampled-token setting.

\paragraph{Sampled-based OPD.}
Recent work has studied sampled OPD from several perspectives, including
teacher--student compatibility, rollout length, sampling, and token-level
optimization.
OPSA \citep{ding2026doesonpolicydistillationreally} observes that
learning under sampled reverse KL is concentrated on low-probability
tokens.
Building on similar observation, SuRe
\citep{shao2026tokenlevelanalysissampledtokenreversekl} analyzes the
underlying gradient structure and further strengthens updates to
low-confidence tokens.
PowerOPD \citep{zhao2026poweropdstabilizingonpolicydistillation}
focuses on a different issue, replacing the unbounded log-ratio reward
with a bounded transformation to improve training stability. Our work revisits the gradient allocation of sampled OPD from the
opposite direction.
We ask whether the weak updates assigned to high-confidence tokens are
actually justified, and show that these tokens can still provide useful
learning signal.
This motivates Bernoulli OPD, which directly uses the bounded
teacher--student probability difference as its training signal.
As a result, high-confidence teacher corrections remain effective
without explicitly favoring either low- or high-confidence tokens.

\paragraph{Binary objectives in computer vision distillation.}
Binary formulations have been explored in
computer vision distillation.
Decoupled Knowledge Distillation (DKD)
\citep{zhao2022decoupledknowledgedistillation} separates the target class from the remaining classes in image classification.
Related binary formulations have also been used to distill class
predictions in object detection \citep{yang2024bridgingcrosstaskprotocolinconsistency}. Bernoulli OPD uses a related binary view in a different setting. For each student-sampled token, we group the vocabulary into the sampled
token and its complement, and match the corresponding teacher and
student Bernoulli probabilities. Our goal is not to separate class-level knowledge, but to preserve useful teacher corrections on high-confidence tokens that receive weak gradients under sampled reverse KL.

\section{Background and Motivation}
\label{sec:background}

Let $x$ denote a prompt and
$y_{1:T}\sim p_{\mathrm S}(\cdot\mid x)$ a trajectory sampled from the
current student.
At step $t$, let $c_t=(x,y_{<t})$ denote the visited context and $y_t$
the sampled next token.
We define the student and teacher probabilities of the sampled token as
\begin{equation}
    \ps = p_{\mathrm S}(y_t\mid c_t),
    \qquad
    \pt = p_{\mathrm T}(y_t\mid c_t).
\end{equation}
The teacher is frozen, and the sampled trajectory is treated as fixed
within each optimization step.

\paragraph{Vanilla sampled reverse KL.}
Define the teacher--student log-probability gap as
\begin{equation}
    A_t = \log \pt - \log \ps.
\end{equation}
A common sampled surrogate for reverse-KL OPD is
\begin{equation}
    \mathcal{L}_{\mathrm{Vanilla},t}
    =
    -\sg(A_t)\log \ps,
    \label{eq:vanilla}
\end{equation}
where $\sg(\cdot)$ denotes stop-gradient.
The commonly used K2 objective,
$\tfrac{1}{2}(\log\pt-\log\ps)^2$,
induces the same gradient with respect to the student parameters.

Let $\zs$ denote the logit of the sampled token.
Its gradient is
\begin{equation}
    \frac{\partial \mathcal{L}_{\mathrm{Vanilla},t}}
         {\partial \zs}
    =
    -A_t(1-\ps),
    \label{eq:vanilla-grad}
\end{equation}
while the $\ell_1$ norm of the corresponding full-vocabulary logit
gradient is
\begin{equation}
    \left\|
    \nabla_{\mathbf z}
    \mathcal{L}_{\mathrm{Vanilla},t}
    \right\|_1
    =
    2|A_t|(1-\ps).
    \label{eq:vanilla-l1}
\end{equation}

These expressions show that vanilla sampled OPD does not depend only on
teacher--student disagreement.
Its gradient contains an additional factor $1-\ps$ from the softmax
geometry.
As the student becomes more confident in the sampled token, this factor
approaches zero and weakens the corresponding optimization signal.
Therefore, even when the teacher still disagrees with a confident
student prediction, the token can receive only a small gradient.

\paragraph{High-confidence tokens dominate rollouts but receive weak gradients.}
This confidence-dependent suppression is substantial in practice.
As shown in Figure~\ref{fig:vanilla-gradient-distribution}, more than
$70\%$ of sampled tokens lie in the highest-confidence region
$\ps\in[0.9,1.0)$.
Despite dominating the rollout distribution, these tokens have much
smaller per-token gradients and consequently account for only a limited
share of the total gradient mass.
Optimization under vanilla sampled OPD is therefore distributed very
differently from the sampled-token distribution itself.

Importantly, this imbalance does not by itself imply that
high-confidence tokens are uninformative.
A small gradient can arise either because the teacher and student already agree, or because the objective suppresses the gradient when the student is highly confident.
Distinguishing these two cases is important for understanding whether the gradient allocation of vanilla OPD is desirable.

\begin{figure}[t]
    \centering
    \includegraphics[width=0.66\linewidth]
    {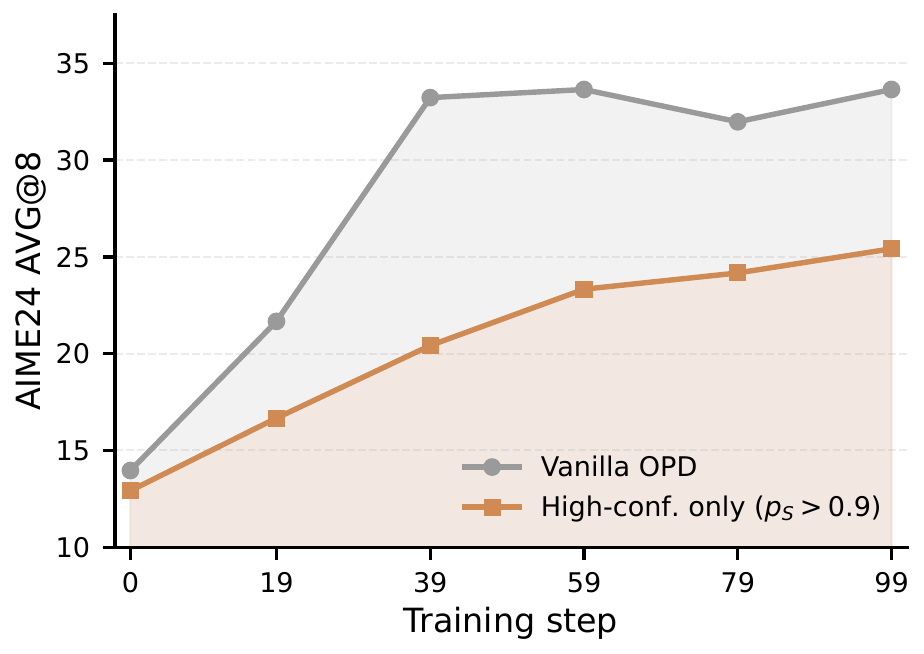}
    \caption{
    \textbf{Training with high-confidence sampled tokens.}
    This diagnostic experiment uses our primary
    Qwen3-4B-Instruct-2507 $\rightarrow$ Qwen3-1.7B setting.
    Vanilla OPD restricted to tokens with $\ps>0.9$ still yields
    substantial performance gains despite the strong gradient
    suppression in this confidence region.
}
    \label{fig:highconf-training}
\end{figure}
\paragraph{High-confidence tokens can still provide useful training signal.}
To test whether the high-confidence region contains useful learning
signal, we train Vanilla OPD using only sampled tokens with
$\ps>0.9$.
Figure~\ref{fig:highconf-training} shows that the student continues to
improve substantially throughout training despite discarding all
lower-confidence tokens.
High-confidence tokens are therefore not devoid of useful teacher
supervision, even though vanilla OPD assigns them relatively little
gradient mass.

Together, these results reveal a mismatch between
token informativeness and
objective-induced gradient allocation.
High-confidence tokens are both abundant and learnable, yet their
updates are systematically attenuated by the confidence dependence of
sampled reverse KL.
This motivates an objective that preserves teacher--student
disagreement as a learning signal without additionally suppressing
updates merely because the student is already confident.

\section{Methods}
\label{sec:method}

\begin{figure*}[t]
    \centering
    \includegraphics[width=\textwidth]
    {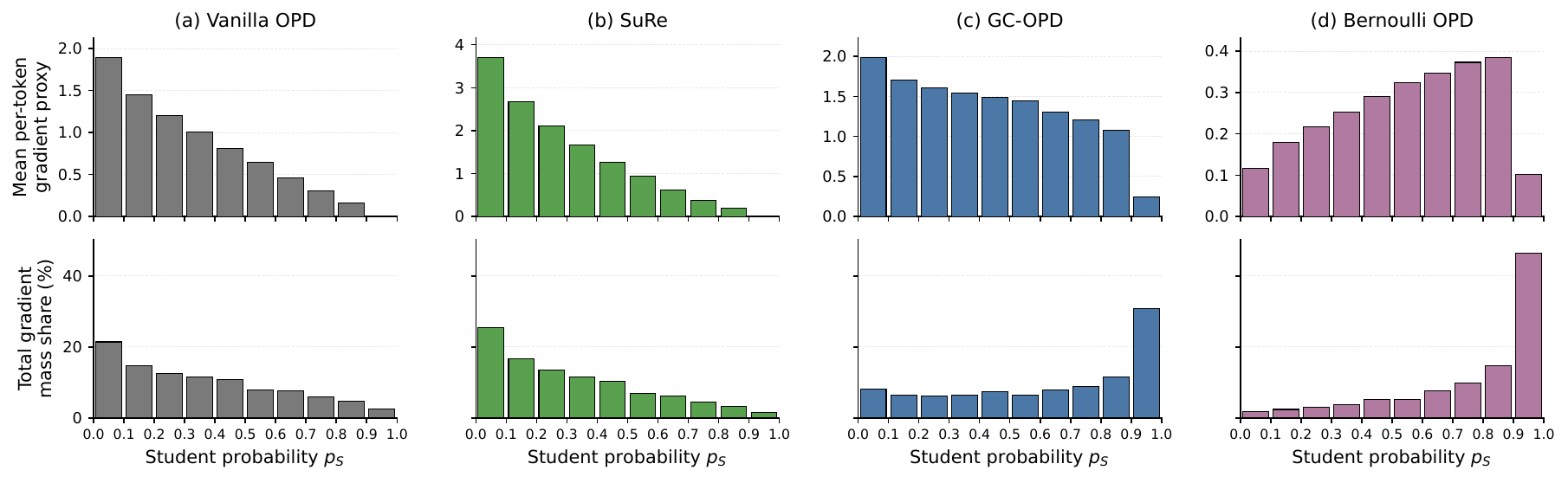}
    \caption{
    \textbf{Gradient allocation under different OPD objectives.}
    All methods are evaluated on the same fixed student rollouts.
    The top row shows the \textbf{mean per-token gradient proxy}
    across student-confidence bins, and the bottom row shows the
    corresponding \textbf{share of total gradient mass}.
    Compared with (a) Vanilla OPD, (b) SuRe further emphasizes
    low-confidence updates, whereas (c) GC-OPD and
    (d) Bernoulli OPD substantially restore gradient contribution
    in high-confidence regions. Top-row y-axis scales differ across objectives; the bottom row reports normalized gradient-mass shares.
    }
    \label{fig:four-loss-gradient}
\end{figure*}

Vanilla sampled reverse-KL OPD combines teacher--student
disagreement with a confidence-dependent factor $1-p_S$,
which strongly suppresses updates on high-confidence sampled tokens.
We study this issue through two complementary objectives.
GC-OPD removes only the explicit confidence-dependent suppression
and serves as a diagnostic control.
Our main method, Bernoulli OPD, replaces the sampled log-ratio
objective with a Bernoulli cross-entropy objective whose
sampled-token logit gradient is bounded.
Figure~\ref{fig:four-loss-gradient} compares their gradient allocation
with Vanilla OPD and SuRe on the same fixed student rollouts.

\subsection{Geometry-Corrected OPD}

We first isolate the explicit $1-\ps$ factor in vanilla OPD while preserving the same teacher--student log-probability signal.
Recall
\begin{equation}
    A_t=\log\pt-\log\ps.
\end{equation}
We replace the sampled-token log probability with its log odds:
\begin{equation}
    \mathcal{L}_{\mathrm{GC},t}
    =
    -\sg(A_t)
    \log\frac{\ps}{1-\ps}.
    \label{eq:gc}
\end{equation}
Because the derivative of the sampled-token log odds with respect to its logit $\zs$ is one,
\begin{equation}
    \frac{\partial \mathcal{L}_{\mathrm{GC},t}}
         {\partial \zs}
    =
    -A_t.
    \label{eq:gc-grad}
\end{equation}

GC-OPD therefore removes the explicit $1-\ps$ suppression of vanilla OPD while retaining the same log-ratio discrepancy.
As shown in Figure~\ref{fig:four-loss-gradient}(c), this substantially restores gradient contribution from high-confidence tokens.
We use GC-OPD primarily as a diagnostic control to test whether the confidence-dependent softmax geometry itself limits useful teacher corrections.

\subsection{Bernoulli OPD}

GC-OPD removes the explicit confidence suppression but still relies on the unbounded log-probability gap.
We therefore introduce \textbf{Bernoulli OPD}, which instead measures teacher--student disagreement directly in probability space.

For each sampled token $y_t$, consider the binary event
\begin{equation}
    B_t=\mathbb{1}[Y=y_t].
\end{equation}
The categorical teacher and student distributions induce Bernoulli marginals with parameters $\pt$ and $\ps$.
We distill these marginals using binary cross-entropy:
\begin{equation}
\begin{split}
    \mathcal{L}_{\mathrm{Bern},t}
    ={}&
    -\pt\log\ps \\
    &-(1-\pt)\log(1-\ps).
\end{split}
\label{eq:bern}
\end{equation}

This objective is a sampled token-versus-complement coarse-graining evaluated at student-visited states, rather than a full-vocabulary forward KL.
Its derivative with respect to the sampled-token logit is
\begin{equation}
    \frac{\partial \mathcal{L}_{\mathrm{Bern},t}}
         {\partial \zs}
    =
    \ps-\pt.
    \label{eq:bern-grad}
\end{equation}

Bernoulli OPD therefore uses the direct teacher--student probability gap.
Unlike vanilla OPD, the correction is not multiplied by an additional vanishing $1-\ps$ factor.
The probability gap is also bounded in $[-1,1]$, avoiding the large scale that can arise from log-probability ratios.
When teacher and student agree, the update naturally vanishes; when they disagree, the correction remains visible even at high student confidence.

This change produces a markedly different gradient allocation.
Figure~\ref{fig:four-loss-gradient}(d) shows that Bernoulli OPD shifts substantial optimization mass toward the high-confidence region that is nearly ignored by vanilla OPD, while still assigning larger per-token updates where teacher--student disagreement is stronger.

\paragraph{Summary.}

\begin{table*}[t]
\caption{
\textbf{Structural comparison of sampled OPD objectives.}
$A=\log p_T-\log p_S$.
We compare the sampled-token gradient together with whether each method
removes the log-probability gap and the explicit $1-p_S$ factor.
GC-OPD serves as a diagnostic variant, while Bernoulli OPD is our main
proposed objective.
}

\label{tab:objectives}
\centering
\small
\setlength{\tabcolsep}{8pt}
\begin{tabular}{lccc}
\toprule
Method
& Sampled-token gradient $\partial L/\partial \zs$
& Remove log gap?
& Remove $1-\ps$ factor? \\
\midrule

Vanilla OPD
& $-A(1-\ps)$
& No
& No \\

SuRe
& $-w_t A(1-\ps)$
& No
& No \\

PowerOPD
& $-(\pt^\alpha-\ps^\alpha)(1-\ps)$
& Yes
& No \\

\midrule
\myrowcolor{BlueLight}
GC-OPD (Ours)
& $-A$
& No
& \textbf{Yes} \\

\myrowcolor{BlueLight}
Bernoulli OPD (Ours)
& $\ps-\pt$
& \textbf{Yes}
& \textbf{Yes} \\

\bottomrule
\end{tabular}
\end{table*}

Table~\ref{tab:objectives} summarizes the sampled-token gradients of
our two objectives together with vanilla OPD.
GC-OPD keeps the original teacher--student log-probability gap while
removing the explicit $1-\ps$ factor, allowing us to isolate the effect
of confidence-dependent gradient suppression.
Bernoulli OPD further replaces the log-ratio signal with the bounded
teacher--student probability difference, and serves as our main
objective.

For a clearer comparison with existing sampled OPD methods, we also
include SuRe and PowerOPD in the same gradient view.
SuRe changes the relative weighting of token updates, while PowerOPD
uses a bounded power transformation of the teacher--student signal.
Together, this comparison shows how our objectives differ from related
methods in both the form of the training signal and its
confidence-dependent gradient behavior.
We further study confidence rebalancing and gradient scale in
Section~\ref{sec:additional-controls}.

\begin{table*}[t]
\centering
\caption{
\textbf{Main results across four teacher settings.}
The first four benchmarks use Avg@32 and MATH500 uses Avg@4. GC-OPD is a diagnostic variant, while Bernoulli OPD is our main proposed objective.
}
\label{tab:main-results}
\small
\setlength{\tabcolsep}{7pt}

\begin{tabular}{lccccccc}
\toprule
\textbf{Method}
& \textbf{AIME24}
& \textbf{AIME25}
& \textbf{HMMT25}
& \textbf{AMC23}
& \textbf{MATH500}
& \textbf{Avg.}
& \textbf{$\Delta$ vs. Vanilla} \\
\midrule

Student: Qwen3-1.7B
& 12.08 & 9.79 & 5.10 & 47.50 & 70.65 & 29.02 & -- \\

\midrule
\multicolumn{8}{l}{
\textbf{Teacher: Qwen3-4B-Instruct-2507}
} \\

Vanilla OPD
& 33.96 & 21.77 & 15.00 & 68.36 & 82.80 & 44.38 & -- \\

SuRe
& 35.10 & 23.02 & 14.48 & 69.84 & 82.30 & 44.95 & +0.57 \\

PowerOPD
& 38.65 & 26.46 & 18.23 & 71.09 & 83.90 & 47.67 & +3.29 \\

\myrowcolor{BlueLight}
GC-OPD
& 38.02 & 27.60 & 18.96 & 75.39 & 85.60 & 49.11 & +4.73 \\

\myrowcolor{BlueLight}
Bernoulli OPD
& \textbf{39.58}
& \textbf{33.02}
& \textbf{19.90}
& \textbf{78.67}
& \textbf{87.65}
& \textbf{51.76}
& \textbf{+7.38} \\

\midrule
\multicolumn{8}{l}{
\textbf{Teacher: Qwen3-8B-DAPO-RL}
} \\

Vanilla OPD
& 50.21
& 39.79
& \textbf{25.42}
& 82.19
& 89.20
& 57.36
& -- \\

SuRe
& 48.75
& 39.58
& \textbf{25.42}
& 82.66
& 88.80
& 57.04
& -0.32 \\

PowerOPD
& \textbf{52.50}
& 39.17
& 24.38
& 85.31
& 89.45
& 58.16
& +0.80 \\

\myrowcolor{BlueLight}
GC-OPD
& \textbf{52.50}
& 40.00
& 24.38
& \textbf{85.94}
& 86.60
& 57.88
& +0.52 \\

\myrowcolor{BlueLight}
Bernoulli OPD
& 51.04
& \textbf{41.25}
& \textbf{25.42}
& 82.97
& \textbf{91.20}
& \textbf{58.38}
& \textbf{+1.02} \\

\midrule
Student: DS-R1-Distill-Qwen-1.5B
& 30.21 & 24.58 & 13.96 & 69.22 & 81.45 & 43.88 & -- \\

\midrule
\multicolumn{8}{l}{
\textbf{Teacher: DS-R1-Distill-Qwen-7B}
} \\

Vanilla OPD
& 28.44
& 23.02
& \textbf{15.10}
& 70.31
& 82.75
& 43.92
& -- \\

SuRe
& 29.06
& 23.02
& 14.17
& 70.31
& 82.90
& 43.89
& -0.03 \\

PowerOPD
& 30.63
& \textbf{23.96}
& 13.75
& 70.16
& 82.60
& 44.22
& +0.30 \\

\myrowcolor{BlueLight}
GC-OPD
& \textbf{31.15}
& 23.33
& 14.27
& 70.16
& 82.75
& 44.33
& +0.41 \\

\myrowcolor{BlueLight}
Bernoulli OPD
& 30.10
& 23.85
& 14.38
& \textbf{72.66}
& \textbf{83.50}
& \textbf{44.90}
& \textbf{+0.98} \\

\midrule
\multicolumn{8}{l}{
\textbf{Teacher: Skywork-OR1-Math-7B}
} \\

Vanilla OPD
& 35.00
& \textbf{27.60}
& 16.46
& 75.78
& 85.40
& 48.05
& -- \\

SuRe
& 33.33
& 27.19
& 16.77
& 73.59
& 83.85
& 46.95
& -1.10 \\

PowerOPD
& 34.90
& 27.40
& 16.46
& 75.94
& 84.80
& 47.90
& -0.15 \\

\myrowcolor{BlueLight}
GC-OPD
& 38.13
& 27.19
& 17.50
& 73.91
& 85.45
& 48.44
& +0.39 \\

\myrowcolor{BlueLight}
Bernoulli OPD
& \textbf{40.42}
& 26.88
& \textbf{17.71}
& \textbf{79.22}
& \textbf{86.40}
& \textbf{50.13}
& \textbf{+2.08} \\

\bottomrule
\end{tabular}
\end{table*}

\section{Experiments}
\label{sec:experiments}

\subsection{Experimental Setup}
\label{sec:setup}

\paragraph{Models and training.}
We evaluate four teacher--student configurations spanning two student
models.
For Qwen3-1.7B, we use Qwen3-4B-Instruct-2507 and
Qwen3-8B-DAPO-RL\footnote{\url{https://huggingface.co/pb09204048/Qwen3-8B-DAPO-iter449-disable-thinking}} as teachers, with
Qwen3-4B-Instruct-2507 as our primary setting
\citep{yang2025qwen3technicalreport}.
For DeepSeek-R1-Distill-Qwen-1.5B, we use
DeepSeek-R1-Distill-Qwen-7B and Skywork-OR1-Math-7B as teachers.
Following the native generation conventions of the corresponding model
families, we evaluate the Qwen3 settings in non-thinking mode and the
DeepSeek-based settings in thinking mode.
Training prompts are drawn from DAPO-Math-17K
\citep{yu2025dapoopensourcellmreinforcement}.
Within each teacher--student setting, all methods share the same student
initialization, rollout prompts, generation mode, and optimization
configuration; only the distillation objective changes, unless otherwise
specified.
Full training and implementation details are provided in
Appendix~\ref{app:implementation}.

\paragraph{Evaluation.}
We evaluate checkpoints on AIME24 \citep{aime_2024} and select the best
checkpoint using Avg@$8$.
Final evaluation covers AIME24, AIME25 \citep{aime_2025},
HMMT25 \citep{balunović2026matharenaevaluatingllmsuncontaminated},
AMC23 \citep{cao2025stepguidedreasoningimproving}, and
MATH500 \citep{lightman2023letsverifystepstep}.
The first four benchmarks are reported with Avg@$32$, while MATH500
uses Avg@$4$.
Decoding parameters and evaluation details are given in
Appendix~\ref{app:implementation}.

\paragraph{Baselines and variants.}
We compare against vanilla sampled reverse-KL OPD,
SuRe \citep{shao2026tokenlevelanalysissampledtokenreversekl}, and
PowerOPD \citep{zhao2026poweropdstabilizingonpolicydistillation}.
We implement SuRe within our framework and evaluate it under the same
unified training setup as our methods.
For PowerOPD, we evaluate both the unified and official reference
configurations and report the better result for each teacher--student
setting to ensure a fair and conservative comparison.
Implementation details and results for both configurations are
provided in Appendix~\ref{app:poweropd-implementation}.

Among our methods, GC-OPD serves as a diagnostic control that removes
the explicit $1-\ps$ suppression while retaining the log-ratio signal,
whereas Bernoulli OPD is our primary proposed objective.
We additionally study Balanced Bernoulli OPD and a scale-matched
Bernoulli control in Section~\ref{sec:additional-controls} to examine
confidence-level rebalancing and the effect of overall optimization
magnitude.

\subsection{Main Results}
\label{sec:main-results}

Table~\ref{tab:main-results} summarizes the main results across four
teacher--student settings.
Bernoulli OPD improves aggregate reasoning performance over vanilla OPD
in all four settings, with gains ranging from $0.98$ to $7.38$ points.
The largest improvement occurs in our primary
Qwen3-4B-Instruct-2507 $\rightarrow$ Qwen3-1.7B setting, where
Bernoulli OPD increases the five-benchmark average from $44.38$ to
$51.76$.
The improvement also persists with the Qwen3-8B-DAPO-RL teacher and
across both settings using the DeepSeek-R1-Distill-Qwen-1.5B student,
indicating that the benefit is not specific to a single student or
teacher family.

The existing baselines show different strengths across settings.
SuRe modifies the confidence preference of sampled OPD, while PowerOPD
uses a bounded transformation of the log-ratio reward and provides a
strong competitive baseline.
Bernoulli OPD achieves the highest aggregate performance across all four
settings.
GC-OPD also improves over vanilla OPD in every setting, providing
evidence that reducing the confidence-dependent suppression of sampled
reverse KL can be beneficial.
Bernoulli OPD further improves over GC-OPD across all four settings.
One possible explanation is that Bernoulli OPD not only removes the
additional suppression of high-confidence tokens, but also expresses
teacher--student disagreement directly through a Bernoulli cross-entropy objective with a bounded sampled-token logit gradient.
This combination may provide a more effective training signal across
different teacher--student settings.

The magnitude of the Bernoulli OPD improvement nevertheless varies
substantially across teachers.
This suggests that correcting gradient allocation determines
\emph{which} teacher corrections can act strongly on the student, but
does not by itself determine how useful those corrections are.
We further analyze this teacher-dependent variation in
Sec.~\ref{sec:teacher-dependent}.

\begin{figure*}[t]
    \centering
    \includegraphics[
        width=0.95\textwidth
    ]{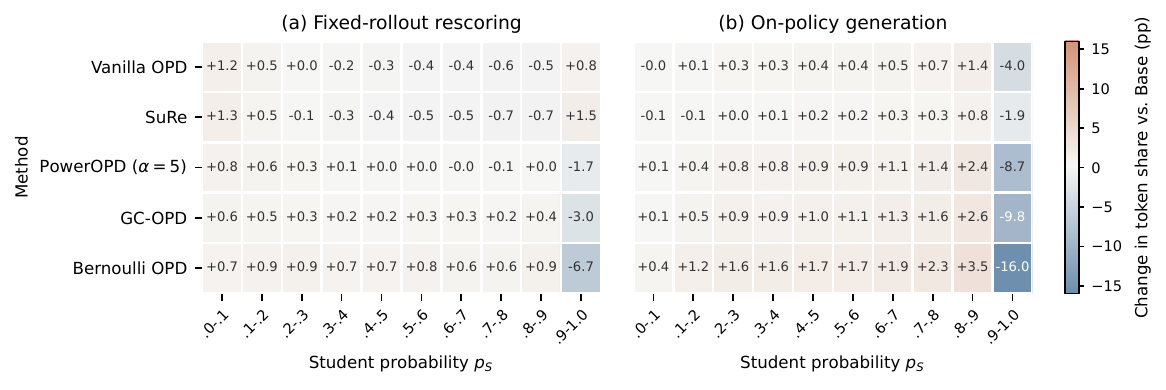}
    \caption{
    \textbf{Confidence redistribution after OPD training.}
    Changes in token share across confidence bins relative to the initial
    student under (a) fixed-rollout rescoring and (b) on-policy generation.
    Bernoulli OPD shows the strongest redistribution away from the
    highest-confidence region.
    }
    \label{fig:confidence-redistribution}
\end{figure*}

\subsection{Additional Variants for Bernoulli OPD}
\label{sec:additional-controls}

We further examine two questions:
\textbf{(1)} does making the gradient allocation more uniform across
confidence regions improve performance?
\textbf{(2)} are the gains of Bernoulli OPD simply due to a larger
overall optimization step?

\paragraph{Balanced Bernoulli OPD.}
Bernoulli OPD substantially restores optimization signal to
high-confidence tokens, but it does not explicitly attempt to equalize
gradient mass across confidence regions.
Its gradient magnitude remains determined by the underlying
teacher--student probability discrepancy, $|p_S-p_T|$, which can itself
vary substantially with student confidence.
This raises a natural question: \emph{once the severe suppression of
high-confidence tokens is removed, would making the gradient allocation
even more balanced across confidence regions lead to further gains?}

To examine this question, we introduce \emph{Balanced Bernoulli OPD} as
a diagnostic extension that explicitly reweights Bernoulli updates
according to the aggregate gradient mass in each confidence region.
Specifically, we partition valid response tokens into $B$ bins according
to their student probability $p_S$.
For each bin $b$, we compute its aggregate Bernoulli gradient mass
\begin{equation}
    M_b
    =
    \sum_{t \in b}
    \left|p_S(t)-p_T(t)\right|.
\end{equation}
Let $\bar{M}$ denote the mean mass across valid bins.
We assign tokens in bin $b$ a detached weight
\begin{equation}
    \tilde{w}_b
    =
    \left(
        \frac{\bar{M}}
        {M_b+\epsilon}
    \right)^{\beta},
\end{equation}
followed by normalization over valid tokens and clipping to limit
extreme weights.
The resulting objective is
\begin{equation}
    \mathcal{L}_{\mathrm{Balanced},t}
    =
    w_{b(t)}
    \mathcal{L}_{\mathrm{Bern},t},
\end{equation}
with sampled-token gradient
\begin{equation}
    \frac{\partial
    \mathcal{L}_{\mathrm{Balanced},t}}
    {\partial z_y}
    =
    w_{b(t)}
    (p_S-p_T),
\end{equation}
since $w_b$ is detached.
We use $B=10$, $\beta=0.5$, and clip the normalized weights to
$[1/5,5]$.

By construction, Balanced Bernoulli pushes the gradient allocation
toward a more even distribution across student-confidence regions than
standard Bernoulli OPD.
If the remaining imbalance under Bernoulli OPD were itself an important
limitation, this stronger rebalancing should provide additional
downstream benefit.
Table~\ref{tab:balanced-bernoulli}, however, shows no such consistent
improvement.
Balanced Bernoulli yields only small gains of $0.11$, $0.27$, and
$0.06$ points in three settings, while decreasing the five-benchmark
average by $0.92$ points with the Skywork-OR1-Math-7B teacher.

\begin{table}[ht]
    \centering
    \caption{
    \textbf{Effect of stronger confidence-level rebalancing.}
    We report the five-benchmark average for standard and Balanced
    Bernoulli OPD across the four teacher--student settings.
    $\Delta$ denotes Balanced Bernoulli minus Bernoulli OPD.
    }
    \label{tab:balanced-bernoulli}
    \small
    \setlength{\tabcolsep}{5pt}
    \begin{tabular}{lccc}
        \toprule
        \textbf{Teacher}
        & \textbf{Bernoulli}
        & \textbf{Balanced}
        & \textbf{$\Delta$} \\
        \midrule

        Qwen3-4B-Instruct
        & 51.76
        & \textbf{51.87}
        & +0.11 \\

        Qwen3-8B-DAPO-RL
        & 58.38
        & \textbf{58.65}
        & +0.27 \\

        DS-R1-Distill-Qwen-7B
        & 44.90
        & \textbf{44.96}
        & +0.06 \\

        Skywork-OR1-Math-7B
        & \textbf{50.13}
        & 49.21
        & -0.92 \\

        \bottomrule
    \end{tabular}
\end{table}

These results clarify that the goal is not to make gradient allocation
uniform across confidence regions.
Rather, the main benefit comes from removing the strong
objective-induced suppression of high-confidence tokens while allowing
the remaining gradient distribution to reflect the natural
teacher--student probability discrepancies present in the data.
The gradient allocation induced directly by Bernoulli OPD therefore
appears sufficient: explicitly flattening it further provides little
additional benefit and can even be detrimental.
We consequently use standard Bernoulli OPD as our primary objective and
treat Balanced Bernoulli OPD only as a mechanism control.

\paragraph{Gradient-scale control.}
Figure~\ref{fig:four-loss-gradient} shows that Bernoulli OPD produces a
substantially different gradient profile from vanilla OPD, raising the
possibility that its gain results simply from a larger overall update.
To isolate this effect, we construct a \emph{scale-matched Bernoulli}
variant whose loss is rescaled at initialization to match the
full-model gradient norm of vanilla OPD.

Table~\ref{tab:scale-control-main} summarizes the result.
Scale matching makes both the initial gradient norm and one-step policy
movement nearly identical to vanilla OPD while preserving Bernoulli
OPD's distinct confidence-dependent gradient allocation.
The downstream advantage also remains: scale-matched Bernoulli reaches
an average accuracy of $52.58$, compared with $44.38$ for vanilla OPD
and $51.76$ for standard Bernoulli OPD.
In this diagnostic setting, the gain therefore cannot be explained solely by a larger global update.
We use scale matching \textbf{only as a diagnostic control} because it requires
calibration against vanilla OPD.
Full calibration details are provided in
Appendix~\ref{app:scale-control}.

\begin{table}[t]
    \centering
    \caption{
    \textbf{Gradient-scale control.}
    Policy move is the mean absolute sampled-token log-probability
    change after one optimizer step; HC mass denotes the gradient-proxy
    mass assigned to $p_S\in[0.9,1.0]$; Avg. is the downstream
    five-benchmark average.
    }
    \label{tab:scale-control-main}
    \small
    \setlength{\tabcolsep}{3.0pt}
    \begin{tabular}{lcccc}
        \toprule
        \textbf{Method}
        & $\mathbf{\|g\|_2}$
        & \textbf{Policy move}
        & \textbf{HC mass}
        & \textbf{Avg.} \\
        \midrule
        Vanilla OPD
        & 8.139
        & 0.00891
        & 1.44\%
        & 44.38 \\

        \myrowcolor{BlueLight}
        Bernoulli OPD
        & 9.993
        & 0.00908
        & 39.42\%
        & 51.76 \\

        \myrowcolor{BlueLight}
        Scale-Matched
        & \textbf{8.138}
        & \textbf{0.00892}
        & 39.42\%
        & \textbf{52.58} \\
        \bottomrule
    \end{tabular}
\end{table}

\subsection{Teacher-Dependent Variation}
\label{sec:teacher-dependent}
The benefit of Bernoulli OPD varies substantially across teacher--student settings. In our primary Qwen3-4B-Instruct-2507 $\rightarrow$ Qwen3-1.7B setting, Bernoulli OPD improves the five-benchmark average over vanilla OPD by $7.38$ points, whereas the gain with Qwen3-8B-DAPO-RL is only $1.02$ points. The two DeepSeek-based settings also show different gains, ranging from
$0.98$ to $2.08$ points. These results suggest that teacher strength or student--teacher performance headroom alone does not determine how much Bernoulli OPD benefits from the teacher signal.

To investigate this difference, we score exactly the same Qwen3-1.7B student rollouts with the 4B-Instruct and 8B-DAPO teachers.
Both exhibit nearly the same confidence-dependent imbalance under vanilla OPD: more than $90\%$ of sampled tokens lie in the highest-confidence region, yet this region receives only about $3\%$ of the vanilla gradient-proxy mass.
Bernoulli OPD restores this contribution to approximately $47$--$49\%$ for both teachers.
Thus, the existence of high-confidence suppression alone is also insufficient to predict the downstream gain.

Our additional analysis suggests that the \emph{quality} of the recovered correction may matter in addition to its magnitude.
The 4B-Instruct teacher provides slightly more outcome-aligned Bernoulli signals on the fixed rollouts, although the statistical difference is not conclusive.
One possible factor is teacher provenance: Qwen3-4B-Instruct-2507 is a newer instruction-tuned checkpoint and may differ substantially in supervised post-training, whereas Qwen3-8B-DAPO-RL is obtained by further RL training a model from the same Qwen3 base generation as the student.
Such differences may affect the compatibility and teachability of teacher supervision on student-generated trajectories.
We provide the detailed signal analysis in Appendix~\ref{app:teacher-dependent}.

\begin{figure}[h]
    \centering
    \includegraphics[width=0.75\columnwidth]
    {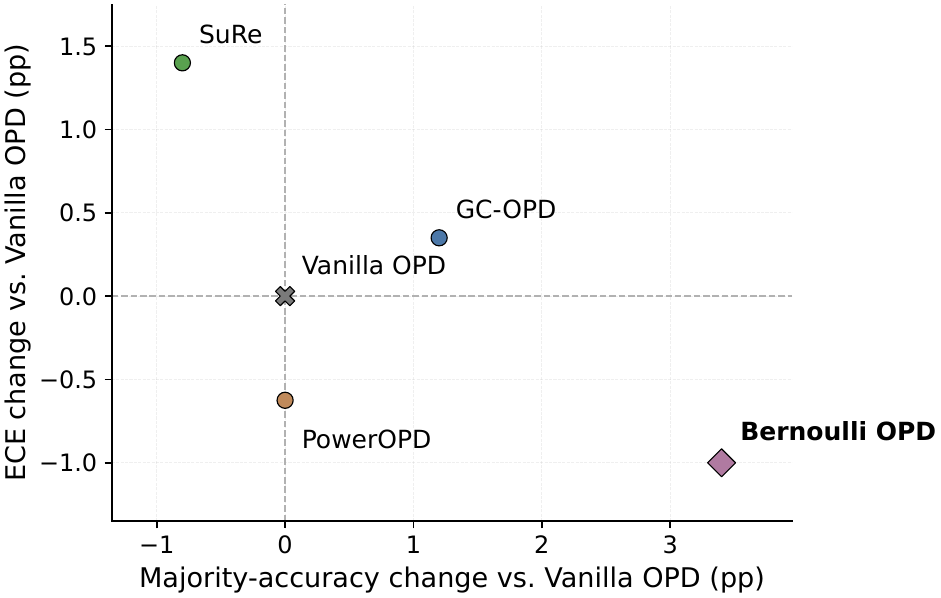}
    \caption{
    \textbf{Answer-level accuracy--calibration trade-off on MATH500.}
Changes in majority-answer accuracy and ECE relative to Vanilla OPD
with eight samples per problem in MATH500.
    }
    \label{fig:calibration-tradeoff}
\end{figure}

\begin{figure*}[t]
    \centering
    \includegraphics[
        width=0.65\textwidth
    ]{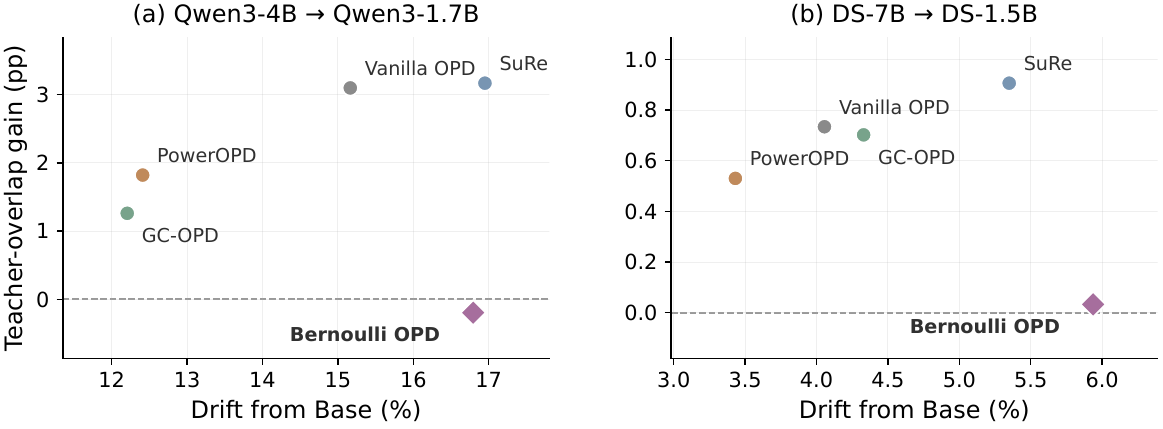}
    \caption{
\textbf{Policy movement after OPD training.}
On fixed student rollouts, we measure policy drift from the initial
student and the corresponding change in Top-20 teacher vocabulary
overlap.
Bernoulli OPD induces substantial policy movement with little or no
increase in global teacher overlap.
For compactness, Qwen3-4B denotes Qwen3-4B-Instruct-2507, while
DS-7B and DS-1.5B denote DeepSeek-R1-Distill-Qwen-7B and
DeepSeek-R1-Distill-Qwen-1.5B, respectively.
}
    \label{fig:policy-movement}
\end{figure*}

\begin{figure*}[t]
    \centering
    \includegraphics[
        width=0.80\textwidth
    ]{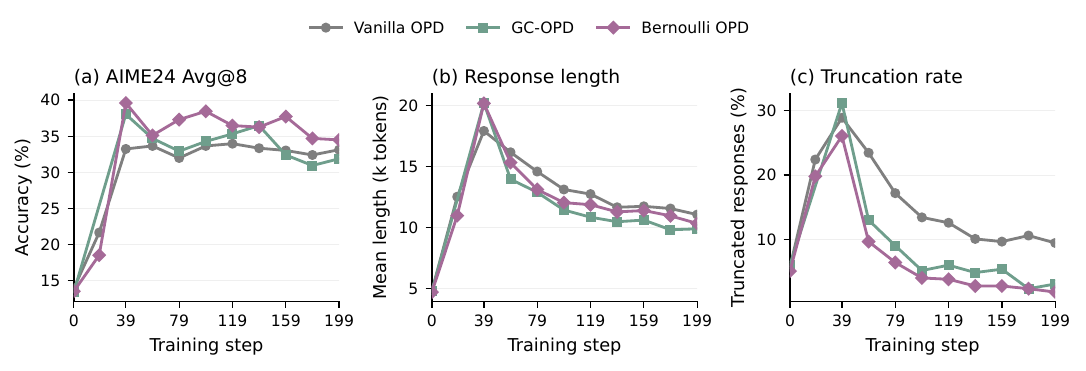}
    \caption{
    \textbf{Training dynamics of different OPD objectives.}
    We report (a) AIME24 Avg@8, (b) mean response length, and
    (c) the fraction of responses truncated at the 32,768-token
    generation limit across training checkpoints.
}
    \label{fig:training-dynamics}
\end{figure*}

\subsection{Model Behaviours}
\label{sec:behavior-analysis}

\paragraph{Confidence redistribution.}

We next examine how different OPD objectives affect the confidence
distribution of the trained student.
Figure~\ref{fig:confidence-redistribution} reports the change in token
share across confidence bins relative to the initial student.

Under fixed-rollout rescoring, where all models are evaluated on the
same prefixes and target tokens, Bernoulli OPD produces the largest
redistribution away from the highest-confidence region.
GC-OPD shows a similar but weaker trend, while Vanilla OPD, SuRe, and
PowerOPD induce substantially smaller changes.
Because the evaluated states and tokens are fixed, this effect cannot be
explained solely by differences in generated trajectories.

The same pattern becomes even more pronounced under on-policy
generation.
Bernoulli OPD substantially reduces the concentration of tokens in the
$p_S\in[0.9,1.0]$ bin and redistributes probability mass toward lower
confidence regions.
Together, these results suggest that the different gradient allocation
of Bernoulli OPD leads to a measurable change in the confidence behavior
of the trained model. Importantly, restoring gradients to high-confidence tokens does not imply increasing their confidence: teacher corrections can instead reduce excessively confident predictions, consistent with the redistribution observed here.

\paragraph{Answer-level calibration.}

We next examine whether the changes in confidence induced by different
objectives translate into meaningful uncertainty at the answer level.
On MATH500, we sample 8 responses per problem and define the
confidence of the predicted answer as the fraction of samples belonging
to its majority answer cluster.
Figure~\ref{fig:calibration-tradeoff} compares each method with Vanilla
OPD in terms of majority-answer accuracy and Expected Calibration Error
(ECE).
Bernoulli OPD exhibits the most favorable point-estimate trade-off,
suggesting that Bernoulli OPD's confidence redistribution
does not come at the expense of answer-level calibration; instead, it
is accompanied by both higher answer consistency accuracy and lower
calibration error.

\paragraph{Policy movement.}

We further examine whether the improvements from Bernoulli OPD can be
explained by globally imitating the teacher's next-token distribution.
Figure~\ref{fig:policy-movement} measures policy movement on fixed Base
rollouts across two teacher--student settings.
We compare the drift from the initial student policy with the
corresponding change in Top-20 vocabulary overlap with the teacher.
Across both settings, Bernoulli OPD substantially changes the student
policy without a comparable increase in global teacher overlap.
In contrast, Vanilla OPD and SuRe exhibit stronger increases in
teacher overlap despite achieving weaker downstream performance.
These results suggest that Bernoulli OPD's gains do not require a corresponding increase in Top-20 vocabulary overlap with the teacher.

\paragraph{Training dynamics.}

We finally examine how accuracy and generation length evolve during
training in our primary Qwen3-4B-Instruct-2507 $\rightarrow$ Qwen3-1.7B setting.
Figure~\ref{fig:training-dynamics} compares Vanilla OPD, GC-OPD, and
Bernoulli OPD on AIME24 using the same evaluation protocol.
All three methods exhibit a sharp but transient increase in response
length early in training, accompanied by a substantial rise in
truncation rate.
Response length and truncation then steadily decrease as training
continues.

Importantly, accuracy does not simply track response length.
Bernoulli OPD maintains higher AIME24 validation accuracy over most later
checkpoints even after its response length and truncation rate have
fallen substantially.
This suggests that its performance gain cannot be explained solely by
generating longer responses, but reflects a different learning
trajectory induced by the objective.

\section{Conclusion}

We show that sampled reverse-KL OPD systematically suppresses updates to
high-confidence tokens, even though these tokens can still provide useful
learning signal.
We introduce Bernoulli OPD, which replaces the log-ratio surrogate with
a Bernoulli cross-entropy objective whose sampled-token logit gradient
directly reflects the bounded teacher--student probability difference.
Across four teacher--student settings, Bernoulli OPD consistently improves over
vanilla OPD.
Our controls further show that the gain is not explained simply by a
larger update or by enforcing more uniform gradients across confidence
regions.
Overall, our results highlight the importance of separating token
informativeness from the gradient geometry of the distillation objective.

\section*{Impact Statement}

This work improves the efficiency of transferring mathematical reasoning
to smaller language models.
While effective distillation can reduce deployment costs, it may also
propagate teacher errors or undesirable behavior.
Our analysis helps interpret such transfer, and all experiments use
public mathematical datasets without sensitive data.

\bibliography{references}
\bibliographystyle{icml2026}

\clearpage
\appendix

\appendix



\section{Additional Implementation Details}
\label{app:implementation}

\subsection{Training Setup}

Unless otherwise specified, the following training settings
apply to our unified experiments. All experiments are conducted on four NVIDIA GH200 GPUs.
Two GPUs are used for actor training, while the remaining two are used
for rollout generation and teacher inference.

Training prompts are sampled from DAPO-Math-17K.
For the Qwen3-1.7B settings, generation is performed in non-thinking
mode.
For the DeepSeek-R1-Distill-Qwen-1.5B settings, we retain the default
thinking behavior of the reasoning models.
For each prompt, we generate one student response at temperature $1.0$
with a maximum generation length of $12{,}000$ tokens.
Training uses a global batch size of $64$ and Adam with a constant
learning rate of $10^{-6}$.

Within each teacher--student setting, all compared methods share the same
student initialization, rollout prompts, generation configuration, and
response masking rules unless otherwise specified.
Each method generates its own on-policy trajectories and obtains
teacher scores on the corresponding student-generated tokens.
The unified experiments additionally share the same loss normalization,
optimizer, and training schedule.
PowerOPD results using its reference training recipe are described
separately in Appendix~\ref{app:poweropd-implementation}.

\subsection{Evaluation and Checkpoint Selection}

We evaluate checkpoints every $20$ training updates on AIME24.
Checkpoint selection uses Avg@$8$ with temperature $0.7$, top-$p$ $0.8$,
top-$k$ $20$, and a maximum response length of $32{,}768$ tokens.
The selected checkpoint is the one with the highest AIME24 Avg@$8$.

Final evaluation uses the same decoding parameters.
We report Avg@$32$ on AIME24, AIME25, HMMT25, and AMC23, and Avg@$4$
on MATH500.
For the Qwen3-8B-DAPO-RL teacher setting, the maximum response length is
increased to $38{,}912$ tokens because the resulting student generations
are substantially longer.
Final-answer accuracy is computed using the same mathematical answer
extraction and grading pipeline across all methods.

\subsection{Balanced Bernoulli OPD}

For Balanced Bernoulli OPD, valid response tokens are partitioned into
$10$ student-confidence bins.
We use $\beta=0.5$ and clip the normalized detached bin weights to
$[1/5,5]$.
The weighting coefficients are computed from aggregate Bernoulli
gradient mass within each confidence bin, as defined in
Section~\ref{sec:additional-controls}.
Balanced Bernoulli OPD is included as an exploratory mechanism control
for studying explicit confidence-level rebalancing rather than as a
fully tuned primary method.

\subsection{PowerOPD Reference Recipe and Controlled Comparison}
\label{app:poweropd-implementation}

We independently implement PowerOPD
\citep{zhao2026poweropdstabilizingonpolicydistillation} within our
training framework.
We first evaluate it under the same unified training setup used by the
other objectives, so that the distillation objective is the main factor
being changed.
We then repeat the experiments using the reference PowerOPD training
recipe.

Table~\ref{tab:poweropd-config} summarizes the main configuration
differences.
We report only settings that differ materially between the two recipes;
both configurations use Adam, gradient clipping of $1.0$, one response
per prompt, DAPO-Math-17K, and $\alpha=5$.
The same four-GH200 hardware allocation is used for both configurations.

\begin{table}[t]
    \centering
    \caption{
    \textbf{Key training differences between the unified and PowerOPD
    reference recipes.}
    Only settings that differ materially between the two configurations
    are shown.
    }
    \label{tab:poweropd-config}
    \small
    \setlength{\tabcolsep}{4.5pt}
    \begin{tabular}{lcc}
        \toprule
        \textbf{Setting}
        & \textbf{Unified}
        & \textbf{Reference} \\
        \midrule
        Optimizer updates
        & 200
        & 1,500 \\

        Learning rate
        & $1\times10^{-6}$
        & $5\times10^{-7}$ \\

        Warmup
        & 0
        & 5\% \\

        Adam $(\beta_1,\beta_2)$
        & $(0.9,\,0.98)$
        & $(0.9,\,0.999)$ \\

        Weight decay
        & 0.1
        & 0 \\

        Global batch size
        & 64
        & 32 \\

        Max.\ rollout length
        & 12,000
        & 1,024 \\

        Sampling $(T,p,k)$
        & $(1.0,\,1.0,\,-1)$
        & $(1.0,\,0.95,\,20)$ \\
        \bottomrule
    \end{tabular}
\end{table}

Table~\ref{tab:poweropd-recipe} compares the resulting downstream
performance.
The reference recipe outperforms the unified configuration in three
of the four teacher--student settings, including our primary
Qwen3-4B-Instruct-2507 setting, while the unified configuration
performs better with Skywork-OR1-Math-7B.
To ensure a competitive and conservative baseline comparison,
we report the higher five-benchmark average from the two configurations
for each teacher--student setting in Table~\ref{tab:main-results}.
This avoids underestimating PowerOPD due to the choice of training
configuration, while the unified results also provide a controlled
comparison under the same optimization setup as our other methods.

\begin{table}[t]
    \centering
    \caption{
    \textbf{PowerOPD under the unified and reference training recipes.}
    We report the five-benchmark average in each teacher--student
    setting.
    $\Delta$ denotes Reference minus Unified.
    Bold values indicate the results used in the main comparison.
    }
    \label{tab:poweropd-recipe}
    \small
    \setlength{\tabcolsep}{5pt}
    \begin{tabular}{lccc}
        \toprule
        \textbf{Teacher}
        & \textbf{Unified}
        & \textbf{Reference}
        & \textbf{$\Delta$} \\
        \midrule
        Qwen3-4B-Instruct-2507
        & 45.42 & \textbf{47.67} & +2.25 \\

        Qwen3-8B-DAPO-RL
        & 57.80 & \textbf{58.16} & +0.36 \\

        DS-R1-Distill-Qwen-7B
        & 44.09 & \textbf{44.22} & +0.13 \\

        Skywork-OR1-Math-7B
        & \textbf{47.90} & 44.24 & -3.66 \\
        \bottomrule
    \end{tabular}
\end{table}

For completeness, the PowerOPD objective used in both configurations is
\begin{equation}
    r_t^{\mathrm{Power}}
    =
    \sg\!\left(
        \pt^{\alpha} - \ps^{\alpha}
    \right),
\end{equation}
with
\begin{equation}
    \mathcal{L}_{\mathrm{Power},t}
    =
    -
    r_t^{\mathrm{Power}}\log\ps,
\end{equation}
where $\alpha=5$ in both configurations.

\subsection{Answer-Level Calibration Details}

For the answer-level calibration analysis in Section~\ref{sec:behavior-analysis}, we sample eight responses per
MATH500 problem.
We extract and normalize the final mathematical answer from each
response and cluster answers by mathematical equivalence using exact
matching together with the MathD and SymPy graders.
Responses for which no valid answer can be extracted are assigned to a
single \texttt{INVALID} cluster.
The largest answer cluster is treated as the model prediction, and its
relative frequency among the eight samples defines the answer-level
confidence.
If multiple clusters are tied for the largest size, we select the
cluster containing the earliest generated response.

We compute Expected Calibration Error (ECE) using 10 equal-width bins
over $[0,1]$:
\begin{equation}
    \mathrm{ECE}
    =
    \sum_{b}
    \frac{n_b}{N}
    \left|
        \mathrm{conf}(b)-\mathrm{acc}(b)
    \right|,
\end{equation}
where $n_b$ is the number of problems in bin $b$, and
$\mathrm{conf}(b)$ and $\mathrm{acc}(b)$ denote its mean answer-level
confidence and empirical accuracy, respectively.
Because each problem has eight samples, the confidence values are
restricted to multiples of $1/8$.
This quantity should be interpreted as sampling-based answer
consistency rather than the model's exact probability assigned to the
correct answer.

\subsection{Numerical stability.}
We compute the Bernoulli objective in FP32 and clamp student and
teacher probabilities to $[\epsilon, 1-\epsilon]$, where
$\epsilon=10^{-6}$. To avoid numerical instability near $p_S=1$,
we evaluate the complementary log-probability using
$\operatorname{log1p}(-\tilde p_S)$, where
$\tilde p_S=\operatorname{clip}(p_S,\epsilon,1-\epsilon)$.
The resulting implementation is
\[
\mathcal{L}_{\mathrm{Bern}}
= -\tilde p_T\log p_S
-(1-\tilde p_T)\operatorname{log1p}(-\tilde p_S),
\]
where $\tilde p_T$ is the clipped, detached teacher probability.
Balanced Bernoulli OPD uses the same stable objective with
an additional detached balancing weight.


\section{Additional Analysis of Teacher-Dependent Gains}
\label{app:teacher-dependent}

\subsection{Controlled Comparison on Fixed Student Rollouts}

To isolate teacher-dependent effects from differences in student
trajectories, we compare Qwen3-4B-Instruct-2507 and Qwen3-8B-DAPO-RL on
exactly the same Qwen3-1.7B student rollouts, comprising
$585{,}751$ sampled tokens.
Because the visited states and sampled tokens are fixed, this comparison
removes rollout variation and changes only the teacher supervision.
Table~\ref{tab:teacher-signal-analysis} summarizes the resulting
token- and sequence-level statistics.

\begin{table*}[t]
    \centering
    \caption{
    \textbf{Teacher signals on identical student rollouts.}
    HC denotes the highest student-confidence region
    ($p_S\in[0.9,1.0]$).
    Bernoulli AUROC measures how well the sequence-level Bernoulli
    teacher signal distinguishes correct from incorrect rollouts.
    }
    \label{tab:teacher-signal-analysis}
    \small
    \setlength{\tabcolsep}{9pt}
    \begin{tabular}{lcccc}
        \toprule
        Teacher
        & Mean $|p_T-p_S|$
        & HC Vanilla Mass
        & HC Bernoulli Mass
        & Bernoulli AUROC \\
        \midrule
        Qwen3-4B-Instruct-2507
        & 0.0597
        & 2.95\%
        & 48.60\%
        & 0.704 \\
        Qwen3-8B-DAPO-RL
        & 0.0535
        & 2.75\%
        & 46.60\%
        & 0.690 \\
        \bottomrule
    \end{tabular}
\end{table*}

The two teachers exhibit similar confidence-dependent gradient
allocation under vanilla OPD.
For both teachers, only about $3\%$ of the gradient-proxy mass is
assigned to the highest-confidence region, whereas Bernoulli OPD
increases this share to approximately $47$--$49\%$.
Nevertheless, the downstream Bernoulli-over-Vanilla gains differ
substantially across the two teacher settings.
Thus, the severity of high-confidence suppression alone is insufficient
to determine the benefit of Bernoulli OPD.

\subsection{Outcome Alignment of Teacher Corrections}

We next ask whether the teacher corrections emphasized by Bernoulli OPD
are related to rollout quality.
We evaluate both teachers on the same 120 fixed Base rollouts from AIME24,
containing 17 correct and 103 incorrect responses.
For each response $i$, we define a sequence-level Bernoulli correction
score as
\begin{equation}
s_i^{(T)}
=
\frac{1}{|\mathcal{M}_i|}
\sum_{t\in\mathcal{M}_i}
\left[
p_T(y_{i,t}\mid c_{i,t})
-
p_S(y_{i,t}\mid c_{i,t})
\right],
\end{equation}
where $\mathcal{M}_i$ contains valid response-token positions after
applying the loss mask and excluding special tokens.
Both teachers are evaluated on the same student-generated tokens using
teacher-forced scoring.
The score averages signed probability differences over valid tokens,
including the observed tokens of truncated responses.

We measure how well this score distinguishes correct from incorrect
responses using AUROC, with rollout correctness as the positive label.
The resulting AUROC is $0.704$ for Qwen3-4B-Instruct-2507 and
$0.690$ for Qwen3-8B-DAPO-RL.
Although the difference is small, the numerically higher AUROC for
Qwen3-4B-Instruct-2507 is consistent with its larger downstream gain.
We do not draw a strong conclusion about teacher differences from
this diagnostic alone.

\subsection{Interpretation}

These results point to two separate factors.
The loss determines how much of the available teacher signal can affect
the student, while the teacher determines how useful that signal is.
Bernoulli OPD reduces the suppression of high-confidence corrections,
but the final gain still depends on the quality of the teacher feedback
on the states visited by the student.

One possible explanation lies in the different model origins and
training histories of the teachers.
Although Qwen3-8B-DAPO-RL undergoes additional reinforcement learning,
its base model, Qwen3-8B, belongs to the same model generation as the
Qwen3-1.7B student.
In contrast, Qwen3-4B-Instruct-2507 comes from a different Qwen3
generation with a potentially different pretraining and post-training history. These differences may affect the nature and usefulness of teacher
corrections on student-generated trajectories.
A full explanation of this teacher-dependent effect is left for future
work.

\section{Additional Details on Gradient-Scale Control}
\label{app:scale-control}

The scale-matched experiment controls for the possibility that
Bernoulli OPD improves simply because it induces a larger overall
optimization step.
Here we provide the construction and calibration details for the control
summarized in Table~\ref{tab:scale-control-main}.

\subsection{Scale-Matched Objective}

To calibrate the scale-matched control, we evaluate Vanilla
and Bernoulli OPD on a shared frozen batch with identical
student initialization, rollouts, teacher scores, and response
masks, using the same optimizer configuration.

On a shared frozen batch, the full-parameter gradient $\ell_2$ norms of
vanilla and Bernoulli OPD are $8.1393$ and $9.9926$, respectively.
We therefore define
\begin{equation}
    c
    =
    \frac{
        \lVert g_{\mathrm{Vanilla}} \rVert_2
    }{
        \lVert g_{\mathrm{Bern}} \rVert_2
    }
    =
    0.8145,
\end{equation}
and optimize
\begin{equation}
    \mathcal{L}_{\mathrm{Matched}}
    =
    c\,\mathcal{L}_{\mathrm{Bern}}.
    \label{eq:matched-bern}
\end{equation}

The coefficient is calibrated once at initialization and remains fixed
throughout training.
Because this is a positive global rescaling, it changes the overall
gradient magnitude while leaving the relative allocation of gradients
across tokens unchanged.

\begin{table*}[t]
\centering
\caption{
\textbf{Robustness across training and evaluation seeds in the primary
Qwen3-4B-Instruct-2507 $\rightarrow$ Qwen3-1.7B setting.}
For Vanilla OPD and Bernoulli OPD, each row corresponds to an
independent training seed and its independently selected checkpoint.
For the initial student, the three rows correspond to independent
evaluation runs of the same model.
Mean and standard deviation are computed across the three runs.
}
\label{tab:seed_robustness}

\small
\setlength{\tabcolsep}{3.5pt}

\begin{tabular*}{\textwidth}{
    @{\extracolsep{\fill}}
    l
    r
    r
    rrrrr
    r
}
\toprule

\textbf{Method}
& \textbf{Seed}
& \textbf{Step}
& \textbf{AIME24}
& \textbf{AIME25}
& \textbf{HMMT25}
& \textbf{AMC23}
& \textbf{MATH500}
& \textbf{Avg.} \\

\midrule

\multicolumn{9}{l}{
    \textit{Initial student: independent evaluation runs}
} \\

Qwen3-1.7B Init.
& 42
& 0
& 12.08
& 9.79
& 5.10
& 47.50
& 70.65
& 29.02 \\

Qwen3-1.7B Init.
& 43
& 0
& 13.96
& 8.54
& 6.67
& 44.53
& 70.60
& 28.86 \\

Qwen3-1.7B Init.
& 44
& 0
& 14.27
& 9.58
& 5.63
& 43.28
& 70.85
& 28.72 \\

\textbf{Mean}
& --
& --
& 13.44
& 9.30
& 5.80
& 45.10
& 70.70
& \textbf{28.87} \\

\textbf{Std.}
& --
& --
& 1.19
& 0.67
& 0.80
& 2.17
& 0.13
& \textbf{0.15} \\

\midrule

\multicolumn{9}{l}{
    \textit{Vanilla OPD: independent training seeds}
} \\

Vanilla OPD
& 1234
& 119
& 33.96
& 21.77
& 15.00
& 68.36
& 82.80
& 44.38 \\

Vanilla OPD
& 5678
& 79
& 35.42
& 20.96
& 15.14
& 67.55
& 82.60
& 44.33 \\

Vanilla OPD
& 13579
& 119
& 34.92
& 22.08
& 15.76
& 68.06
& 82.30
& 44.62 \\

\textbf{Mean}
& --
& --
& 34.77
& 21.60
& 15.30
& 67.99
& 82.57
& \textbf{44.44} \\

\textbf{Std.}
& --
& --
& 0.74
& 0.58
& 0.40
& 0.41
& 0.25
& \textbf{0.16} \\

\midrule

\multicolumn{9}{l}{
    \textit{Bernoulli OPD: independent training seeds}
} \\

Bernoulli OPD
& 1234
& 39
& 39.58
& 33.02
& 19.90
& 78.67
& 87.65
& 51.76 \\

Bernoulli OPD
& 5678
& 39
& 40.73
& 33.33
& 18.54
& 78.91
& 86.95
& 51.69 \\

Bernoulli OPD
& 13579
& 39
& 40.21
& 34.48
& 18.96
& 76.48
& 86.95
& 51.42 \\

\textbf{Mean}
& --
& --
& 40.17
& 33.61
& 19.13
& 78.02
& 87.18
& \textbf{51.62} \\

\textbf{Std.}
& --
& --
& 0.58
& 0.77
& 0.70
& 1.34
& 0.40
& \textbf{0.18} \\

\bottomrule
\end{tabular*}
\end{table*}

\subsection{One-Step Optimization Calibration}

Matching the raw gradient norm does not necessarily imply a matched
optimizer update, particularly under Adam.
We therefore additionally measure the one-step change in the student
policy after applying the optimizer update.

Specifically, starting from the same student parameters, we apply one
optimizer step and measure the mean absolute change in sampled-token
log probability,
\begin{equation}
    M
    =
    \mathbb{E}_t
    \left[
        \left|
        \log p_{\theta^{+}}(y_t\mid c_t)
        -
        \log p_{\theta}(y_t\mid c_t)
        \right|
    \right],
\end{equation}
where $\theta$ and $\theta^{+}$ denote the parameters before and after
the update.

As reported in Table~\ref{tab:scale-control-main}, scale matching makes
both the full-model gradient norm and the resulting one-step policy
movement nearly identical to vanilla OPD.
At the same time, the high-confidence region retains the substantially
larger share of gradient mass induced by the Bernoulli objective.
Thus, the control matches global optimization magnitude without
removing the confidence-dependent redistribution that distinguishes
Bernoulli OPD.

\subsection{Training the Scale-Matched Variant}

We train the scale-matched variant in the primary
Qwen3-4B-Instruct-2507 $\rightarrow$ Qwen3-1.7B setting.
The coefficient $c=0.8145$ is fixed throughout training, with no
per-step normalization or other change to the optimization
configuration.
Checkpoint selection follows the same AIME24 Avg@$8$ protocol as the
main experiments.

Table~\ref{tab:scale-control-main} shows that matching the global update
scale does not remove the downstream advantage of Bernoulli OPD.
This rules out a larger initial gradient norm or larger one-step policy
movement as the sole explanation for the observed improvement.

\subsection{Interpretation}

The scale-controlled experiment separates the overall magnitude of
optimization from its allocation across tokens.
After matching vanilla OPD's global update scale, Bernoulli OPD
preserves both its distinct confidence-dependent gradient allocation and
its downstream advantage.
This supports the interpretation that Bernoulli OPD changes
\emph{where} optimization is allocated rather than merely increasing
\emph{how much} optimization is applied.

Scale-matched Bernoulli slightly outperforms standard Bernoulli OPD in
this run.
We do not interpret this difference as evidence that scale matching is
systematically beneficial, since the experiment uses a single training
seed and requires an additional calibration step.
We therefore treat scale matching strictly as a mechanism control rather
than as a proposed training method.

\section{Robustness Across Training Seeds}
\label{app:seed_robustness}

\subsection{Experimental Protocol}

We additionally evaluate the robustness of our primary comparison to
training randomness.
In the Qwen3-4B-Instruct-2507 $\rightarrow$ Qwen3-1.7B setting, we
repeat both Vanilla OPD and Bernoulli OPD with three independent
training seeds.
For reference, we also evaluate the initial Qwen3-1.7B student three
times using independent stochastic decoding runs.

For every trained model, checkpoint selection is performed independently
using the same AIME24 Avg@$8$ protocol as in the main experiments.
The first four final-evaluation benchmarks use Avg@32, while MATH500
uses Avg@4.

\subsection{Results}

Table~\ref{tab:seed_robustness} shows that both Vanilla OPD and
Bernoulli OPD exhibit small run-to-run variation.
The five-benchmark average of Vanilla OPD is
$44.44 \pm 0.16$, whereas Bernoulli OPD achieves
$51.62 \pm 0.18$.
The approximately seven-point improvement of Bernoulli OPD is therefore
substantially larger than the observed training-seed variation and is
reproduced consistently across all three independent training runs.

The individual benchmark results are also broadly consistent across
training seeds.
In addition, final evaluation averages multiple stochastic generations
per problem, which reduces variance from stochastic decoding.
Given the computational cost of repeating every combination of method
and teacher, and the small training-seed variability observed in this
controlled primary comparison, we report a single training run for the
remaining experiments unless otherwise stated.

\end{document}